\documentclass[letterpaper]{article} 
\usepackage{aaai25}  
\usepackage{times}  
\usepackage{helvet}  
\usepackage{courier}  
\usepackage[hyphens]{url}  
\usepackage{graphicx} 
\usepackage{natbib}  
\usepackage{caption} 
\usepackage{algorithm}

\usepackage{amsmath,amsfonts,bm}

\newcommand{\xmark}{\ding{55}}
\newcommand{\cmark}{\ding{51}}

\def\eqref#1{equation~\ref{#1}}

\def\1{\bm{1}}

\def\vf{{\bm{f}}}

\def\vk{{\bm{k}}}

\def\vq{{\bm{q}}}

\def\vt{{\bm{t}}}

\def\vv{{\bm{v}}}
\def\vw{{\bm{w}}}

\def\mD{{\bm{D}}}

\def\mF{{\bm{F}}}

\def\mP{{\bm{P}}}

\def\mV{{\bm{V}}}

\DeclareMathAlphabet{\mathsfit}{\encodingdefault}{\sfdefault}{m}{sl}
\SetMathAlphabet{\mathsfit}{bold}{\encodingdefault}{\sfdefault}{bx}{n}

\def\gL{{\mathcal{L}}}

\def\sR{{\mathbb{R}}}

\usepackage{graphicx}
\usepackage{url}
\usepackage{multirow}
\usepackage{makecell}
\usepackage{booktabs}
\usepackage{caption}
\usepackage{xcolor}
\usepackage{colortbl}
\usepackage{float}
\usepackage{subfig}
\usepackage{pifont}

\usepackage{pifont}
\usepackage{amsmath}
\usepackage{siunitx} 
\usepackage{algpseudocode}
\definecolor{citecolor}{RGB}{34,139,34}
\definecolor{demphcolor}{gray}{.5}

\newfloat{figtab}{htb}{fgtb}
\makeatletter
  \newcommand\figcaption{\def\@captype{figure}\caption}
  \newcommand\tabcaption{\def\@captype{table}\caption}
\makeatother

\usepackage{newfloat}
\usepackage{listings}
\DeclareCaptionStyle{ruled}{labelfont=normalfont,labelsep=colon,strut=off} 
\floatstyle{ruled}
\newfloat{listing}{tb}{lst}{}
\floatname{listing}{Listing}
\title{Learning to Prompt Your Domain for Vision-Language Models}
\author{
    Guoyizhe Wei\textsuperscript{\rm 1},
    Feng Wang \textsuperscript{\rm 1},
    Anshul Shah\textsuperscript{\rm 1},
    Rama Chellappa\textsuperscript{\rm 1} 
}
\affiliations{
    \textsuperscript{\rm 1}Johns Hopkins University\\

}

\usepackage{bibentry}

\begin{document}

\maketitle
\begin{abstract}

Prompt learning has recently become a very efficient transfer learning paradigm for Contrastive Language Image Pretraining (CLIP) models. Compared with fine-tuning the entire encoder, prompt learning can obtain highly competitive results by optimizing only a small number of parameters, which presents considerably exciting benefits for federated learning applications that prioritizes communication efficiency. However, in this work, we identify that directly transferring prompt learning approaches into federated learning does not yield favorable results since the model often suffers from considerable domain gaps across different clients. To address this issue, we propose ADAPT, a novel domain-aware prompt learning approach that facilitates both intra- and inter-domain prompts across federated participants. The basic idea of ADAPT is that the prompted CLIP should detect the input image's domain correspondence and before making the prediction of its category. Extensive experiments of ADAPT demonstrate its significant efficiency and effectiveness in federated learning. For example, by learning and sharing only 0.08M parameters, our ADAPT attains a 68.4\% average accuracy over six domains in the DomainNet dataset, which improves the original CLIP by a large margin of 14.8\%.

\end{abstract}

\begin{figure*}[t]
    \centering
    \vspace{-5pt}
    \includegraphics[width=0.95\textwidth]{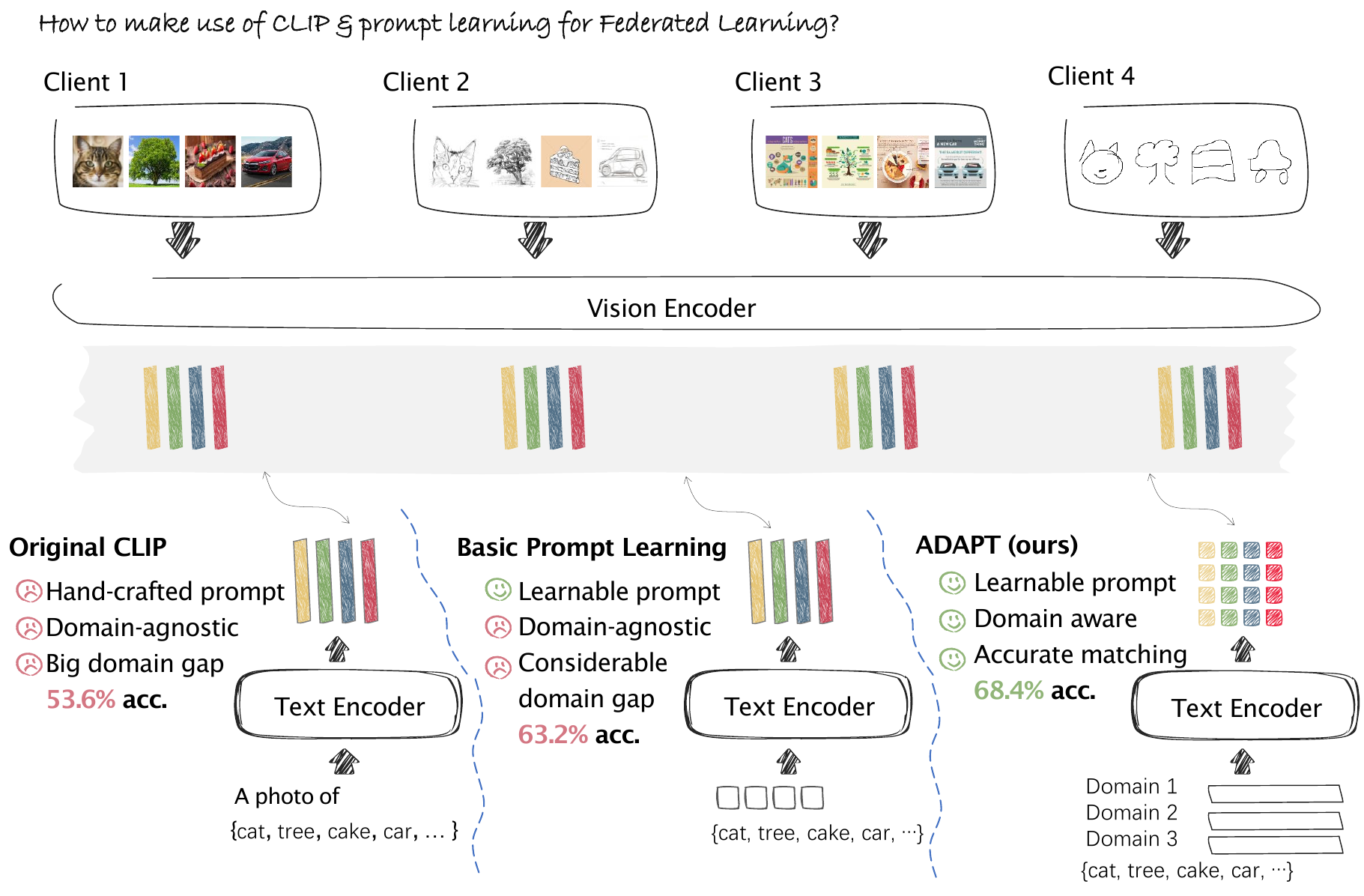}
    \caption{An intuitive comparison of our ADAPT to zero-shot CLIP and the basic domain-agnostic prompt learning approach.}
    \label{fig:over}
    \vspace{-10pt}
\end{figure*}

%

\section{Introduction}\label{Sec:introduction}

Contrastive Language Image Pretraining (CLIP)~\cite{clip} has recently been proven to be a powerful model for multi-modal representation learning. By connecting the feature spaces of images and texts, it enables convenient open-vocabulary classification simply by matching the visual representations of images with the textual embeddings of their class names. Building upon CLIP, the prompt learning technique, which freezes all encoders while introducing learnable tokens at the input side of the model, can help the model adapt to downstream domains with minimal cost.

Compared to the traditional finetuning paradigm, prompt learning techniques built on CLIP can achieve highly competitive results with a minimal amount of learnable parameters (e.g., 0.1\% of encoder parameters). This significant advantage in parameter efficiency motivates us to explore the application of prompt-based CLIP in federated learning --- In federated learning, participants need to frequently share and update the model's learnable parameters, leading to high communication costs and slow convergence rates, while prompt learning offers a highly parameter-efficient approach to fundamentally address these issues.

In this work, we consider a challenging yet realistic federated learning scenario: the participants aim to deal with the same machine learning problem (e.g., image classification with the same target categories), yet their local data originate from different domains. Following prior practice~\cite{fada}, we formulate this scenario using domain-aware datasets like DomainNet~\cite{domainnet}, where there are labeled images sourced from six distinct domains with quite different styles such as real-world, paining, and sketch. Due to the large diversity in input, conventional domain-agnostic federated learning approaches often struggle to generalize well in this problem.

It is noteworthy that this federated learning scenario is more aligned with real-world conditions, as the data heterogeneity among different participants often manifests as variations in feature distributions and image styles, rather than the imbalanced label distributions that most empirical studies use for simulating a non-independent and identical distribution (non-IID). However, in this scenario, due to significant domain gaps between participants, directly applying conventional prompt learning methods cannot yield satisfactory results. We present an intuitive comparison between our method and the existing approaches in Figure~\ref{fig:over}

To address this issue, we propose a simple but highly effective approach called Feder\textbf{A}ted \textbf{D}omain-\textbf{A}ware \textbf{P}rompt \textbf{T}uning (ADAPT). In detail, our ADAPT approach specifically set up a visual and a textual prompt for each potential domain. Each text prompt consists of several learnable tokens that represent textual descriptions indicative of each domain's style information. Each visual prompt denotes a single learnable token that is appended to patch-embedded image tokens. We optimize the prompts through two loss functions: 1) a basic object classification loss, which is applied between the global feature of the input image and the textual representation of its corresponding class name; 2) a domain correspondence loss, which is applied between each pair of visual and textual prompts' output. A detailed framework of our method is illustrated in Figure~\ref{fig:local}

The inference process of ADAPT involves two steps. First, given an input image, the vision encoder can determine the probability or confidence level that the image belongs to each given domain by measuring the attention scores to each visual prompt token. Next, based on these domain correspondence probabilities, we perform cross attention on the representation vectors of the given class name under each textual prompt to produce its final text feature. By finding the text feature that best matches the image feature, we can determine the classification result of the image.

Following extensive theoretical analyses and empirical evaluations, we have identified several significant advantages of ADAPT, which serve as our main contributions:

\begin{itemize}
    \item \textbf{Domain-aware prompt learning}. We assign domain-specific textual prompts to each federated participant, enabling the model to make predictions by input images' corresponding domain information, which effectively addresses the widespread issue of domain gaps in federated learning. Our experimental results demonstrate a significant performance improvement with this design: based on a pretrained CLIP model equipped with a ViT-Base image encoder, our ADAPT method achieves an average accuracy of 68.4\% on DomainNet, which significantly outperforms the zero-shot CLIP's 53.6\%, the basic prompt learning's 63.2\%, and PedProx's~\cite{fedprox} 55.3\% with a same image encoder.

    \item \textbf{Efficient communication}. Our ADAPT approach requires training and sharing only a small fraction of parameters, significantly reducing the communication overhead in federated learning. For instance, in our domain-aware federated learning experiments, ADAPT, with just 0.08M trainable parameters, achieved state-of-the-art performance on the DomainNet~\cite{domainnet}, OfficeHome~\cite{officehome}, and PACS~\cite{pacs} datasets.

    \item \textbf{Superior privacy preservation}. Due to the minimal amount of trainable parameters in ADAPT, traditional federated learning attack algorithms~\cite{deepleak,fedgrad} struggle to reconstruct the local data of participants from model gradients. Additionally, the learnable prompts themselves do not leak customer privacy---we have attempted to decode our prompts but found it difficult to extract any interpretable information from them.
\end{itemize}

\section{Related Work}

\textbf{Federated learning} was first introduced in the Federated Averaging (FedAvg) paper~\cite{fedavg}, addressing machine learning problems with massively distributed private data. To enhance the learning potential of FedAvg, FedProx~\cite{fedprox} adds a $\ell_2$ regularization term into the original federated learning objective. Following FedAvg's success, many follow-up works improve federated learning in terms of privacy-preserving potentials~\cite{fedprivacy1,fedprivacy2}, robustness to heterogeneous data~\cite{scaffold,fedconv}, communication efficiency~\cite{fedcomm1,fedcomm2}, and compatibility to model architectures~\cite{fedgbm,fedvit}. In contrast to general federated learning methods that simulate non-i.i.d. data by partitioning datasets in the label space, many recent works consider federated learning in a more realistic context of domain adaptation~\cite{yao2022federated,Shenaj2023LADD,fada}. Recently, based on advances in multi-modal contrastive learning~\cite{clip}, various works develop CLIP-based federated learning methods. For example, FedCLIP~\cite{fedclip} uses a pre-trained CLIP model and performs federated training on an additional adaptor layer, and PromptFL~\cite{promptfl} proposes to use prompt learning methods for federated optimization.

\textbf{Vision-language models.} Following the success of contrastive pre-training in visual modality~\cite{moco,simclr,byol,dino,simsiam,mocov3}, multi-modal contrastive pre-training has become a common paradigm in recent years as well. A representative work is CLIP~\cite{clip}, which jointly pre-trains a visual and a textual encoder using an InfoNCE objective~\cite{infonce} with around 400 million curated image-text pairs. ALIGN~\cite{align} improves CLIP by scaling up the training dataset to 1.8 billion noisy image-text pairs, and BASIC~\cite{basic} further increases the scale of both data and model. As a result, such CLIP-like models allow zero-shot inference when it comes to transfer learning on downstream tasks.

\textbf{Prompt tuning.} While fine-tuning a pre-trained model for downstream machine learning tasks has traditionally dominated the field of transfer learning, recent progress in prompt learning offers a compelling alternative. Specifically, the prompt tuning techniques fine-tune learnable prompt tokens attached to CLIP's inputs instead of training the entire model~\cite{coop,cocoop,defo,kgcoop}. There also exist prompt tuning protocols for visual modality~\cite{vpt} and both visual and textual modalities~\cite{cpt,upt}. Similarly, there are adapter-based methods designed for CLIP-like models, which also freeze the encoders and only fine-tune several newly attached layers on top of them~\cite{clipadapter,tipadapter}.

\section{Preliminaries}

\subsection{Contrastive Language-Image Pre-training (CLIP)}
CLIP is a weakly supervised learning paradigm that combines visual and language encoders to solve image recognition problems. Formally, CLIP has an image encoder $\mF_V:\sR^{3\times w\times h}\rightarrow \sR^d$ where $w$ and $h$ denotes the input image's spatial resolution and $d$ denotes the dimension of the latent space, and a text encoder $\mF_T: \sR^{l\times d_e}\rightarrow\sR^d$ where $l$ is the length of input sentence and $d_e$ is the dimension of word embedding. CLIP is trained by image-text pairs, in which the text briefly describes the information in the image. By encoding both image and text into the same latent space, CLIP can learn an alignment between visual and textual input with a contrastive loss~\cite{infonce}. During inference, CLIP supports zero-shot classification by match the visual representation of input image and the textual representation of target class names.



\subsection{Prompt Tuning for Vision and Language}

Despite CLIP's impressive zero-shot inference capabilities, there remains a noticeable accuracy gap in comparison to in-domain fine-tuning. However, fine-tuning the CLIP model may easily break the well-established alignment between vision and language, and CLIP will therefore loses the ability of open-vocabulary inference. Instead, prompt tuning attaches learnable tokens to the input, leaving the feature encoders fixed, which allows the model to retain its zero-shot and open-set inference abilities while significantly improves in-domain accuracy.

\textbf{Textual Prompt Tuning (TPT)}. As previously mentioned, CLIP's text query consists of a hand-crafted prompt (also referred to as prefix) such as ``A photo of a'' and a class name such as ``dog''. TPT replaces the prefix by learnable vectors~\cite{coop}. During training, both CLIP's vision and language encoders are frozen and only the prompt vectors are optimized.


\textbf{Visual Prompt Tuning (VPT)}. The prompt tuning protocol also works for visual input if the image encoder is a transformer-like model such as the Vision Transformer~\cite{vit}. Specifically, this method attaches trainable vectors to the patch-wise embedded image, and uses an additional head to project the output. In VPT, only the prompt tokens and the head are optimized.



\begin{figure*}[t]
    \centering
    \includegraphics[width=0.8\textwidth]{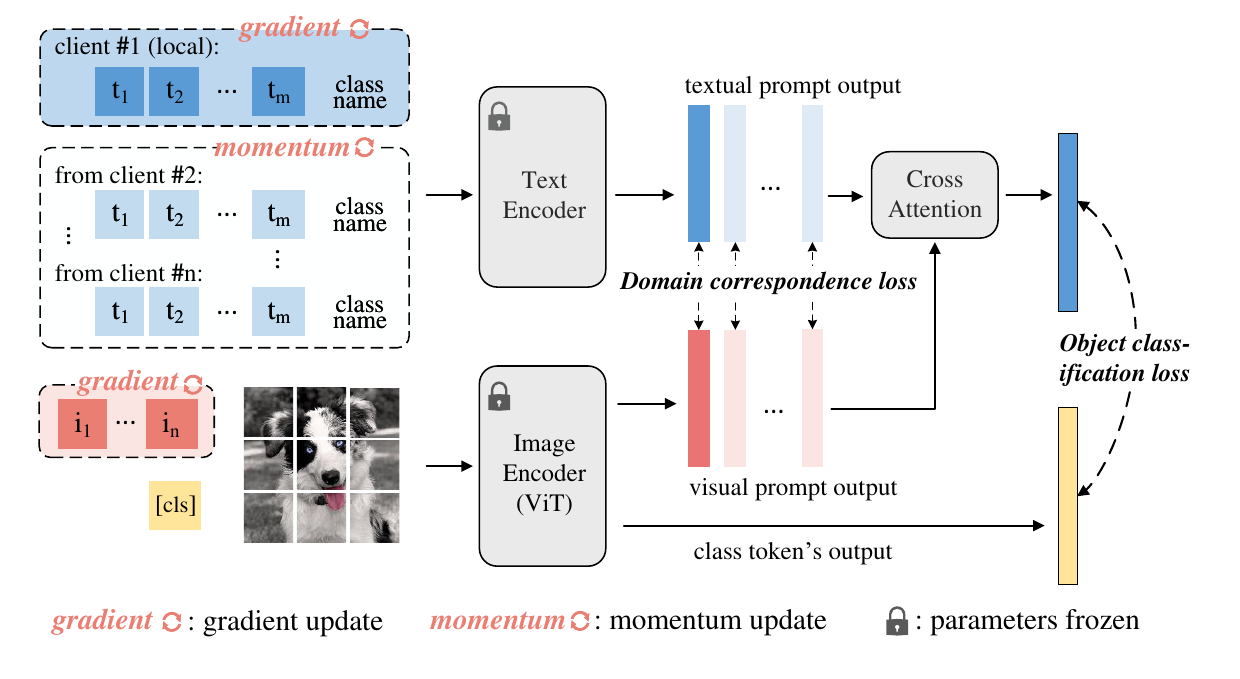}
    \caption{Local training framework. We load a pre-trained CLIP model and freeze both its image and text encoders. For each client, we feed the text encoder with n text prompts followed by class names, where one is optimized by the gradients and the rest $n-1$ are loaded from other clients with momentum update. We feed the image encoder with n learnable prompt tokens followed by patch-wise embedded images, where the prompt tokens are optimized by gradients.}
    \label{fig:local}
    \vspace{-10pt}
\end{figure*}

\section{Methodology}\label{sec:feddpt}




\textbf{Problem Formulation.} Supposing there are $n$ clients that desire to deal with the same machine learning problem, e.g., image classification with the same target categories. The $n$ clients possess their own training data that originate from $n$ distinct domains. In other words, each client stands for a specific domain. We simulate this scenario using domain adaptation datasets like DomainNet~\cite{domainnet}, which encompass images from six different domains including clipart,
information graph, painting, quickdraw, real-world images, and sketch. As the image features exhibit significant variation across different domains, it is indeed a challenging task for federated optimization. However, it is a realistic scenario because many times, the data heterogeneity between clients arises from differences in feature distributions rather than label distributions. Notably, our setting is compatible with the task that clients have non-i.i.d. labels. In our ablation study, we also further divide each domain into five splits with non-i.i.d. categories.

\subsection{Local training}

With CLIP, a very simple way to deal with domain shift is to use domain-aware prompt contexts for text queries. For example, in DomianNet, when we use prefix ``a painting of a'' for the painting domain, and use ``a sketch of a'' for the sketch domain, the predictions can be more accurate and robust. This idea is also referred to as domain-specific prompts~\cite{dapl}, while employing learnable text prompts can further improve the predictive performance. Inspired by this observation, we propose to use domain-specific prompts for CLIP's text encoder. Formally, we define a text prompt by a sequence of learnable tokens:
\begin{equation}
    \mP_T = [\vt]_1[\vt]_2\dots[\vt]_m \in \sR^{m\times d_e},
\end{equation}

where $m$ is the length of prompt and each token $[\vt]_i\in\sR^{d_e}$ has the same dimension as CLIP's word embedding.

Figure~\ref{fig:local} illustrate our ADAPT's local training framework and process. We initialize ADAPT by loading the same CLIP model for each client and freezing the parameters of both the image encoder $\mF_V$ and the text encoder $\mF_T$. For our task, we have $n$ text prompts $\mP_T^1,\mP_T^2,\dots,\mP_T^n$ corresponding to the $n$ domains. During local training, the $n$ text prompts are shared among the clients, yet the $i$-th prompt $\mP_T^i$ can only be trained by the $i$-th client (we will detail this mechanism later). We separately feed the encoder $\mF_T$ with all the $n$ text prompts followed by a class name, leading to $n$ representation vectors $\vf_T^1,\vf_T^2,\dots,\vf_T^n$, where
\begin{equation}
    \vf_T^i=\mF_T(\mP_T^i,[\mbox{class name}]).
\end{equation}
Note that we suppose each $\vf_T^i$ stands for the representation of the class name in the $i$-th domain.

We define visual prompts by $n$ learnable tokens $[\vv]_1,[\vv]_2,\dots,[\vv]_n$ which also correspond to the $n$ domains. During local training, we feed the visual encoder $\mF_V$ (ViT architecture) with a class token [cls] (directly loaded from CLIP), $n$ visual prompts, and the patch-wise embedded image, leading to an image representation vector
\begin{equation}
    \vf_V = \mF_V([\mbox{cls}], [\vv]_1,[\vv]_2,\dots,[\vv]_n, [\mbox{image}]).
\end{equation}

We obtain the final textual representation through a cross attention layer. To minimize the number of parameters as much as possible, here we replace the query, key, and value projection matrices in the cross attention block with identity matrices, which we empirically find does not affect performance too much. Formally, denoting $\vq_{\mbox{cls}}$ as the query vector of the class token, and $\vk_i$ as the key vector of the $i$-th prompt token in $\mF_V$'s last self-attention block, we have $\vw=[w_1,w_2,\dots,w_n]$ with
\begin{equation}\label{eq:weight}
    \vw_i = \frac{\exp(<\vq_{\mbox{cls}}, \vk_i>/\tau_d)}{\sum_j\exp(<\vq_{\mbox{cls}}, \vk_j>/\tau_d)},
\end{equation}
where $\tau_d$ is a temperature coefficient. We regard each component $w_i$ as the visual feature's correlation to the $i$-th domain, and compute the final text output by
\begin{equation}\label{eq:ft}
    \vf_T = \sum_{i=1}^n w_i\vf_T^i.
\end{equation}

During training, we optimize the model (actually learnable parameters only appear in prompts) by an object classification loss, which is a cross-entropy function applied between $\vf_V$ and $\vf_T$, and a domain correspondence loss which is another cross entropy function applied between each pair of visual and textual outputs. Here we explain why we optimize these parameters. We desire the $i$-th text prompt $\mP_T^i$ to represent the features of the $i$-th domain in the latent space of textual embeddings. However, the $i$-th client only possesses images from the $i$-th domain, so we cannot train $\mP_T^j\ (j\neq i)$ yet instead load them from other clients. We introduce visual prompts to detect the correlations between an input image and the $n$ domains, so it is fine to optimize all of them. A detailed comparison of different training strategies can be found in our ablation study (see Table~\ref{tab:abl-tpu} and~\ref{tab:abl-vpu} for details).

\subsection{Parameters Aggregation}

As mentioned above, for the $i$-th client, we optimize $\mP_T^i$ by gradients and load $\mP_T^j\ (j\neq i)$ from other clients, so the aggregation of text prompts does not involve parameter merging processes (e.g. averaging).

Suppose there is a centralized parameter server --- although ADAPT also works for decentralized communication --- and the clients upload their corresponded text prompt to it in each communication round. The server concatenates the $n$ uploaded text prompts and sends to every client. For visual parameters, as all visual prompts are optimized by every client, we perform federated averaging in the server and then send the merged parameters to each client. Note that we do not need to share CLIP encoders' parameters as each client is initialized with the same CLIP model and its parameters are frozen during training.

This parameter aggregation paradigm works well for ADAPT, yet may create a minor problem for the text encoder. Specifically, after each communication round, the external text prompts of the $i$-th client, i.e., $\mP_T^j\ (j\neq i)$ will be re-loaded. We observe that this sudden change of parameters often negatively affects our model. To address this issue, we propose to apply momentum update (also referred to as exponential moving average) to the external text prompts. Formally, we have
\begin{equation}\label{eq:mom}
    [\vt]^s = \alpha[\vt]^{s-1} + (1-\alpha)[\vt],
\end{equation}
where $[\vt]^s$, $[\vt]^{s-1}$ denote the prompt tokens at the $s$ and $s-1$ step, and $[\vt]$ denotes the vector received from other clients, and $\alpha\in[0, 1]$ is a coefficient to control the smoothness. The details of our ablation study related to momentum update can be found in Table~\ref{tab:abl-tpu}.

\subsection{Privacy Preservation}

\begin{figure}
    \centering
    \includegraphics[width=0.45\textwidth]{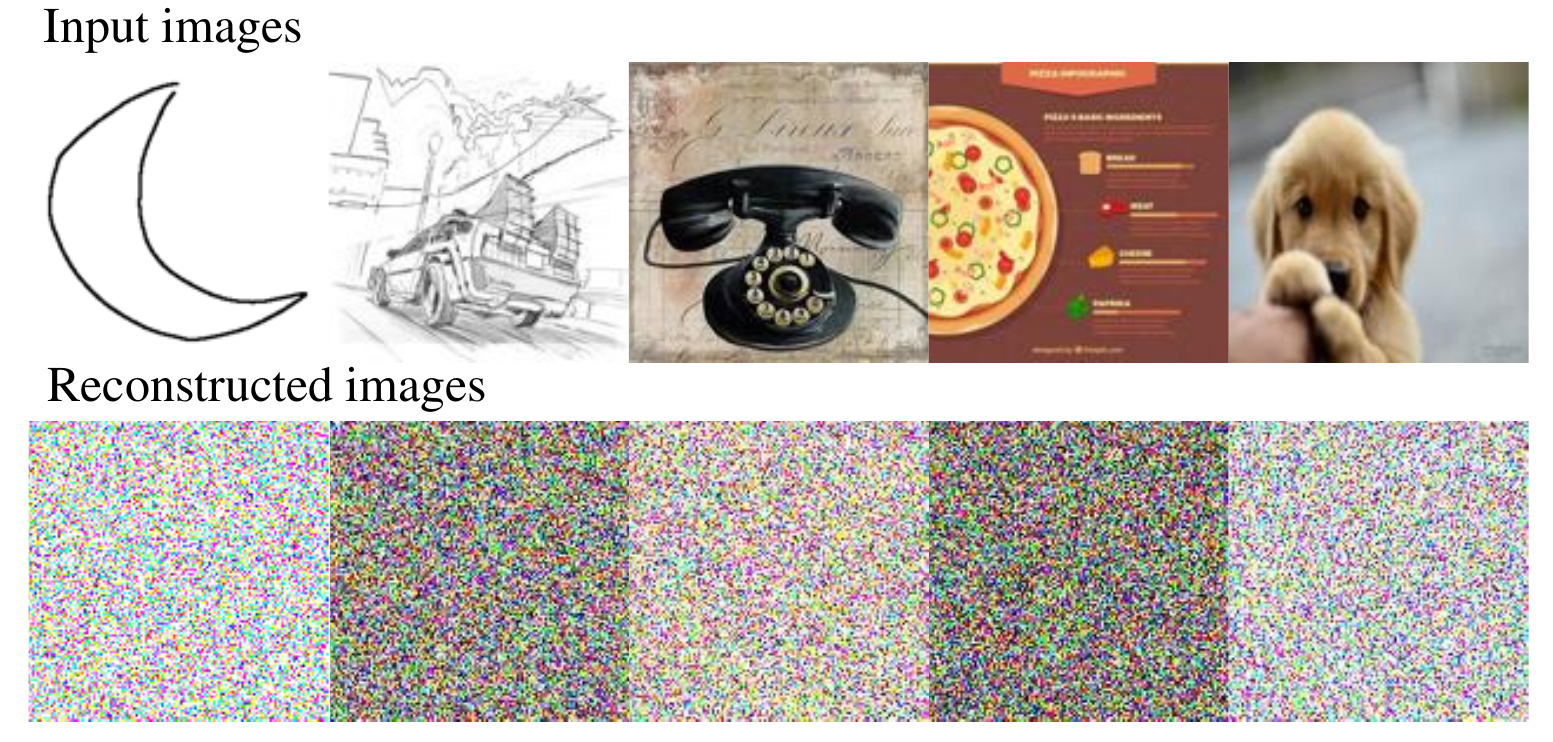}
    \caption{Examples of reconstructed images produced by \textit{Deep Leakage from Gradient}~\cite{deepleak}. It shows that such gradient attack algorithms cannot reconstruction meaningful information from our model.}
    \label{fig:rec}
    \vspace{-10pt}
\end{figure}

In ADAPT, there are two potential ways to expose participants‘ private data. First, similar to most federated learning algorithms, our ADAPT requires to share gradient information across all participants, so some private information might be able to be reconstructed by gradient-based attacking algorithms such as \textit{Deep Leakage from Gradient} (DLG)~\cite{deepleak}. However, as ADAPT only introduces a minimal number of learnable parameters, these attacking algorithms cannot extract sufficient information from gradients to reconstruct participants' local data, which gives our ADAPT a significant privacy advantage over traditional federated learning algorithms. Figure~\ref{fig:rec} presents examples of DLG for our model, where there does not appear any meaningful information corresponding to the input.

Another potential way to expose privacy is decoding the trained text prompts, which might contain some statistical information of participants. However, our experiments showcase that this is difficult as well. we follow CoOp~\cite{coop} to decode each text prompt by finding a standard vocabulary word with minimum Euclidean distance to it in the embedding space, and summarize the interpretation results for DomainNet in Table~\ref{tab:word}. It shows that our prompts tend to capture some high-level and abstract semantics that are difficult to be summarized to standard natural words. 

\begin{table}[htbp]
\centering
\caption{Nearest Words of textual prompts learned by ADAPT in DomainNet dataset. N/A means non-Latin characters. It shows that our prompts tend to capture high-level and abstract semantics that are difficult to summarize using standard natural language words found in the dictionary.}
\label{tab:word}
\begin{tabular}{p{0.05\linewidth}>{\centering\arraybackslash}p{0.1\linewidth} >{\centering\arraybackslash}p{0.1\linewidth} >{\centering\arraybackslash}p{0.1\linewidth} >{\centering\arraybackslash}p{0.1\linewidth}>{\centering\arraybackslash}p{0.1\linewidth}>{\centering\arraybackslash}p{0.1\linewidth}>{\centering\arraybackslash}p{0.1\linewidth}}
\toprule
\# & {clipart} & {info} & {paint} & {quick} & {real} & {sketch} \\ 
\midrule
1 & \textasciitilde  & fe   & N/A  & N/A    & \textdegree  & kd \\
2 &   N/A      &  \# & dng        & ,          & ...    & with     \\
3 & lh        & bh      & some   & ?           & N/A    & N/A   \\
4 & and         & N/A   & lh        & N/A       & the    & pjf   \\
\bottomrule
\end{tabular}
\vspace{-10pt}
\end{table}





\section{Experiments}



\textbf{Datasets and Baselines}. We evaluate our ADAPT and baseline methods on three domain adaptation image classification benchmarks: the DomainNet~\cite{domainnet}, OfficeHome~\cite{officehome}, and PACS~\cite{pacs} datasets, with details can be found in Appendix. We first consider the baselines of CLIP and its adapted models to federated learning. The \underline{\textit{Zero-shot CLIP}}, which infers by aligning images to class names with a hand-crafted prompt, is a direct baseline to evaluate whether in-domain tuning is necessary for vision-language models in federated learning. We also introduce \underline{\textit{Single-domain tuning}}, which applies textual prompt tuning~\cite{coop} to CLIP only in the local domain, as another baseline to testify whether it is helpful to combine the information across multiple domains. There are also domain-agnostic federated learning approaches based on CLIP such as \underline{\textit{PromptFL}}~\cite{promptfl}, \underline{\textit{pFedPG}}~\cite{pFedPG}, \underline{\textit{FedAPT}}~\cite{fedapt} and \underline{\textit{FedCLIP}}~\cite{fedclip}, which train text prompt and an adapter layer in federated learning fashion, respectively. To further validate the effectiveness of our method, we also compare it with conventional federated learning algorithms \underline{\textit{FedAvg}}~\cite{fedavg} and \underline{\textit{FedProx}}~\cite{fedprox} that are not based on CLIP. We equip these two baselines by a 50-layer ResNet~\cite{resnet} and a base-scale vision transformer with 16$\times$16 patch size~\cite{vit}, both being pre-trained on ImageNet-1k~\cite{imagenet}.

\begin{table*}[ht!]
    \centering
    \caption{Test accuracy (\%) on \textbf{DomainNet}. The \textit{info g.}, \textit{paint.}, and \textit{quick d.} denote the domains of \textit{infogragh}, \textit{painting}, and \textit{quickdraw}, respectively. Our results are marked in \colorbox{cyan!10}{blue}. The best results in each domain are \textbf{bolded}.}
    
    \begin{tabular}{>
    {\raggedright\arraybackslash}p{0.36\linewidth} >
    {\centering\arraybackslash}p{0.05\linewidth} >
    {\centering\arraybackslash}p{0.065\linewidth} >
    {\centering\arraybackslash}p{0.05\linewidth} >
    {\centering\arraybackslash}p{0.08\linewidth} >
    {\centering\arraybackslash}p{0.05\linewidth} >
    {\centering\arraybackslash}p{0.05\linewidth} >
    {\centering\arraybackslash}p{0.05\linewidth}}
    \toprule
    \multirow{2}{*}{\textbf{Method}} & \multicolumn{7}{c}{\bf DomainNet}   \\
    \cmidrule(lr){2-8}
    & clipart   & info g.  & paint.  & quick d.  & real  & sketch  & avg.\\\midrule
    
    Zero-Shot CLIP~\cite{clip}& 66.1 & 40.6 & 62.3 & 13.5  & 80.4 & 58.5 & 53.6\\
    Single-Domain Tuning & 72.3 & 47.2 & 67.1 & 18.8 & 83.6 & 65.8 & 59.1 \\\midrule
    
    \multicolumn{8}{l}{\textit{\textcolor{demphcolor}{Conventional federated learning methods:}}} \\
    FedAvg (\textit{ResNet-50 }) & 40.2 & 61.1 & 57.6 & 33.5 & 75.6 & 60.3 & 54.7\\
    FedAvg (\textit{ViT-B/16 }) & 42.4 & 60.7 & 57.0  & 30.4 & 79.8 & 61.1 & 55.2\\
    FedProx (\textit{ResNet-50 })~\cite{fedprox} & 41.5 & 62.0 & 56.8 & 34.9 & 79.2 & 62.6 & 56.2\\
    FedProx (\textit{ViT-B/16})~\cite{fedprox} & 40.5 & \textbf{63.1} & 57.4 & 29.7 & 81.2 & 59.8 & 55.3\\\midrule
    \multicolumn{8}{l}{\textit{\textcolor{demphcolor}{Domain-agnostic vision-language tuning methods:}}} \\
     PromptFL~\cite{promptfl} & 76.0 & 50.2 & 70.4 & 33.5 & 81.2 & 67.8 & 63.2\\
     FedCLIP~\cite{fedclip} & 74.1 & 48.3 & 68.5 & 31.8 & 80.5 & 58.6 & 60.3 \\
     pFedPG~\cite{pFedPG} & 73.9 & 49.2 & 69.8 & 32.2 & 81.4 & 62.6 & 61.5\\
     FedAPT~\cite{fedapt} & 76.3 & 49.8 & 69.2 & 35.7 & 81.5 & 68.2 & 63.5\\
     \midrule\rowcolor{cyan!5}
    
    ADAPT (ours) & \textbf{77.5}   & \textbf{63.1}  &\textbf{70.5}  &\textbf{41.6}  &\textbf{85.7}  &\textbf{72.1}  &\textbf{68.4}\\
    \bottomrule
    
    \end{tabular}
    \vspace{-5pt}
    \label{tab:main1}
\end{table*}

\textbf{Implementation details}. For our ADAPT, we employ a pre-trained CLIP model with a ViT-Base/16 image encoder, so each textual and visual prompt token has the dimension of 512 and 768, respectively. We set the length of each textual prompt sequence $m=16$ for better robustness, which follows the practice of TPT~\cite{coop}. By default, the number of clients is determined by the number of domains for each dataset, i.e. $n=6$ for DomainNet and $n=4$ for OfficeHome and PACS. We train both our model and the baseline models for 200 epochs and execute the aggregation or broadcast process after every one epoch. We train the ResNet-based models and prompt tokens by a SGD optimizer with 0.01 learning rate, 0.9 momentum, and 0.005 weight decay. ADAPT instead uses AdamW~\cite{adamw} optimizer with $\beta_1=0.9$, $\beta_2=0.999$, 5e-4 learning rate, and 0.01 weight decay for transformer-based models. We set the temperature coefficient $\tau_d=0.1$ in Equation~\ref{eq:weight}, and set the momentum update ratio $\alpha=0.99$ in Equation~\ref{eq:mom}. If not specified, all reported results are average numbers over three trials.

\subsection{Main Results} \label{Sec:main results}

Table~\ref{tab:main1} shows that our ADAPT significantly outperforms baseline methods on DomainNet, with notably high improvement in the ``quickdraw" domain at 41.6\% accuracy. This underscores the effectiveness of our prompt learning approach, which requires fewer trainable parameters, enhancing robustness even with larger models. In contrast, traditional methods like FedAvg and FedProx show minimal or negative gains, especially when upgrading from ResNet-50 to ViT-Base. Our ADAPT also achieves higher average accuracy and lower standard deviation compared to domain-agnostic methods, demonstrating better resilience against domain shifts. We further evaluate the models on OfficeHome and PACS, with the results are summarized in Table~\ref{tab:main2}. The experiments on these benchmarks also support our conclusion of ADAPT's effectiveness by demonstrating higher average accuracy and lower deviation across domains. Specifically, we improve the zero-shot CLIP by 4.3\% average accuracy and 0.3\% standard deviation over four domains in OfficeHome. We also observe that overall, the prompt-based methods consistently outperform the conventional federated learning algorithms that require to train the entire model. This confirms the benefits of employing parameter-efficient approaches in federated learning, and explains why we choose to use prompt tuning to address the domain shift issues.

 \vspace{-4pt}
\begin{table*}[thp!]
    \centering
    \caption{Test accuracy (\%) on \textbf{OfficeHome} and \textbf{PACS}. Domains include \textit{art}, \textit{clipart}, \textit{product}, and \textit{real-world} for OfficeHome, and \textit{photo}, \textit{art painting}, \textit{cartoon}, and \textit{sketch} for PACS. Our results are marked in \colorbox{cyan!10}{blue}. The best results are \textbf{bolded}.}
    
    \begin{tabular}{l >
    {\centering\arraybackslash}p{0.045\linewidth} >{\centering\arraybackslash}p{0.045\linewidth} >{\centering\arraybackslash}p{0.045\linewidth} >{\centering\arraybackslash}p{0.045\linewidth} >{\centering\arraybackslash}p{0.045\linewidth} >{\centering\arraybackslash}p{0.045\linewidth} >{\centering\arraybackslash}p{0.045\linewidth} >{\centering\arraybackslash}p{0.045\linewidth} >{\centering\arraybackslash}p{0.045\linewidth} >{\centering\arraybackslash}p{0.045\linewidth}}
    \toprule
    \multirow{2}{*}{\textbf{Method}}   & \multicolumn{5}{c}{\bf OfficeHome}  & \multicolumn{5}{c}{\bf PACS} \\
    \cmidrule(lr){2-6}\cmidrule(lr){7-11}
     & Ar & Cl & Pr & Rw & Avg. & P & A & C & S & Avg.\\
    \midrule
     Zero-Shot CLIP& 79.5 & 63.1 & 85.3 & 86.5 & 78.6 & 99.8 & 96.9 & 98.8 & 87.7 & 95.8 \\
    Single-Domain & 80.0 & 65.2 & 87.5 & 86.9 & 79.9 & 99.8 & 97.2 & 99.1 & 88.9 & 96.3\\\midrule
    \multicolumn{8}{l}{\textit{\textcolor{demphcolor}{Conventional federated learning methods:}}} \\
    FedAvg (\textit{ResNet-50}) & 66.3 & 49.4 & 77.1 & 77.9 & 67.7 & 89.6 & 52.5 & 78.6 & 76.1 & 74.2 \\
    FedAvg (\textit{ViT-B/16}) & 67.9 & 49.6 & 77.5 & 81.0 & 69.0 & 91.3 & 54.8 & 79.2 & 77.9 & 75.8 \\
    FedProx (\textit{ResNet-50})& 68.8 & 50.5 & 78.6 & 80.3 & 69.6 & 91.7 & 57.0 & 81.8 & 80.2 & 77.7 \\
    FedProx (\textit{ViT-B/16}) & 70.4 & 51.3 & 80.3 & 82.4 & 71.1 & 92.0 & 59.4 & 83.5 & 81.6 & 79.1\\\midrule
    \multicolumn{8}{l}{\textit{\textcolor{demphcolor}{Domain-agnostic vision-language tuning methods:}}} \\
    PromptFL~\cite{promptfl} & 79.8 & 65.6 & 89.5 & 89.1 & 81.0 & 99.9 & 97.1 & 99.0 & 90.6 & 96.7 \\
    FedCLIP~\cite{fedclip} & 79.1 & 65.0 & 88.6 & 88.4 & 80.3 & 99.8 & 97.4 & 98.9 & 89.0 & 96.3 \\\midrule
    \rowcolor{cyan!5}
    Ours &\textbf{82.6} & \textbf{68.2} & \textbf{90.5} & \textbf{90.3} & \textbf{82.9} & \textbf{99.9} & \textbf{98.0}  &\textbf{99.1}  &\textbf{91.7}  & \textbf{97.2}\\
    \bottomrule
    
    \end{tabular}
    \vspace{-10pt}
    \label{tab:main2} 
\end{table*}


\subsection{Ablation Studies}\label{sec:abl}

We first dissect ADAPT model to ablate its performance gains. ADAPT comprises two primary components: visual prompts and domain-specific text prompts. By dissecting these components, we get three more variants of our method: 1) \textit{Visual Only}, it leverages learnable prompt tokens for only image input and uses CLIP's hand-crafted prompt for texts. 2) \textit{Textual Only}, it discards the visual prompt tokens of ADAPT and uses learnable text prompts only. Note that in the absence of visual prompts, we cannot get the weight $w_i$ (see Equation~\ref{eq:weight} and ~\ref{eq:ft}) for each domain, so the text prompts from external clients should also be discarded. We instead aggregate the textual prompts by federated averaging~\cite{fedavg}. 3) \textit{Domain-Agnostic}, it retains both ADAPT's visual and textual prompts but decouples them, i.e., we do not perform the weighted sum process in Equation~\ref{eq:ft}, which can be considered as a simple combination of the modes \textit{Textual Only} and \textit{Visual Only}.

We summarize the results in Table~\ref{tab:abl-cpn}. Since we introduce visual prompt tuning for combining domain information rather than enhancing the visual feature extraction abilities, we do not attach an additional head for the image encoder as in~\cite{vpt}. Therefore, the \textit{Visual Only} mode cannot yield significant performance improvements. We also observe that tuning textual prompts results in a 5.5\% increase in accuracy, and when tuning them in a federated learning fashion, we achieve an additional 4.1\% improvement (\textit{Textual Only}). Notably, compared to the simple visual-and-textual prompt tuning with 63.5\% accuracy, ADAPT achieves a much higher result of 68.4\%, demonstrating the crucial significance of our domain-aware design.

\begin{table}[hbt!]
    \centering
    \caption{Ablation study to model components. We report the average accuracy (\%) over six domains in DomainNet. \textit{VPT} and \textit{TPT} denote whether using visual or textual prompts.}
    \vspace{-5pt}
    
    \begin{tabular}{lccccc}
    \toprule
    \bf Method & Fed. & VPT & TPT & domain & acc. \\\midrule
    Zero-Shot CLIP & \xmark & \xmark & \xmark & \xmark & 53.6\\
    Single-Domain & \xmark & \xmark & \cmark & \xmark & 59.1 \\
    Visual Only & \cmark & \cmark & \xmark & \xmark & 54.2 \\
    Textual Only & \cmark & \xmark & \cmark & \xmark & 63.2 \\
    Domain-Agnostic & \cmark & \cmark & \cmark & \xmark & 63.5 \\
    \rowcolor{cyan!5}
    ADAPT & \cmark & \cmark & \cmark & \cmark & 68.4\\
    \bottomrule
    \end{tabular}
    \label{tab:abl-cpn}
    \vspace{-15pt}
\end{table}

\textbf{Momentum update, prompt length, and communication frequency.} We consider three more factors that may affect results. As mentioned in Section~\ref{sec:feddpt}, we update the external text prompts by exponential moving average to prevent parameters' sudden change. Table~\ref{tab:abl-tpu} presents comparisons regarding the update mechanism for text prompts, where the accuracy drops by 2.2\% in the absence of momentum update. If we train all text prompt tokens in every client, i.e., we disregard the relationship between text prompts and domains, the accuracy drops by 4.4\% as it makes ADAPT a domain-agnostic approach.

By default, we aggregate the visual prompt tokens by federated averaging, as separately training each token in a specific domain does not yield better performance (see Table~\ref{tab:abl-vpu}). As shown in Table~\ref{tab:abl-pl}, we set each textual prompt length to $m=16$, as it works more robust than a shorter prompt ($m=4$), and when we further increase the length, the model tends to overfit and accuracy drops. In Table~\ref{tab:abl-com} we also assess the impact of communication frequency by varying it to 0.5, 1, and 2 training epochs per communication round. It shows that compared to our default setup of one epoch per communication round, more frequent aggregation (0.5 epoch/round) does not lead to improved performance, while conversely, infrequent communication (2 epochs/round) results in a 0.5\% accuracy degradation.

\begin{table}[h!]
    \centering
    \caption{Ablation studies. We report the average accuracy over six domains in DomainNet. The \textit{mtm.} denotes momentum update. Our default setup is marked in \colorbox{cyan!10}{blue}. The best results of each ablation study is \bf bolded.}\label{tab:abl-4-all}
    \hspace{+4pt}
    \subfloat[Text prompt update.\label{tab:abl-tpu}]{
    \begin{tabular}{p{0.25\linewidth}c}
        \toprule
        \bf  Mode & acc. \\\midrule\rowcolor{cyan!5}
        w/ mtm. & \bf 68.4 \\
        w/o mtm. & 66.2 \\
        train all & 64.0 \\
        \bottomrule
    \end{tabular}
    }
    \subfloat[Visual prompt update.\label{tab:abl-vpu}]{
    \begin{tabular}{p{0.25\linewidth}c}
        \toprule
         \bf  Mode & acc. \\\midrule\rowcolor{cyan!5}
         average & \bf 68.4 \\
         split w/ mtm. & 68.3 \\
         split w/o mtm. & 67.5 \\\bottomrule
    \end{tabular}
    }
    \hspace{+2pt}
    \subfloat[Prompt length.\label{tab:abl-pl}]{
    \begin{tabular}{p{0.25\linewidth}c}
        \toprule
         \bf \#tokens & acc. \\\midrule
         4 & 67.5 \\\rowcolor{cyan!5}
         16 & \bf 68.4 \\
         32 & 68.0 \\\bottomrule
    \end{tabular}
    }
    \subfloat[Comm. frequency\label{tab:abl-com}]{
    \begin{tabular}{p{0.2\linewidth}c}
        \toprule
        \bf \#eps/round & acc. \\\midrule
        0.5 & \bf 68.4 \\\rowcolor{cyan!5}
        1 & \bf 68.4 \\
        2 & 67.9 \\\bottomrule
    \end{tabular}
    } 
\vspace{-25pt}
\end{table}

\section{Conclusion}
This work introduces ADAPT, a novel federated learning approach explicitly designed to address the key challenges of domain shift and communication efficiency. Our method strategically combines CLIP and prompt learning techniques for both visual and textual inputs, thereby enhancing parameter-efficiency and minimizing communication costs, while maintaining robustness in federated optimization involving heterogeneous data. Furthermore, we confront the pervasive issue of domain shift across clients by introducing domain-specific prompts and facilitating correlations between visual and textual representations through self-attention mechanisms. These innovations result in a domain-aware federated learning methodology that consistently demonstrates outstanding effectiveness. Notably, our experiments reveal a remarkable achievement—an average accuracy of 68.4\% across six domains in the DomainNet dataset, marking an impressive 14.8\% improvement over the original CLIP model. In comparisons with traditional federated learning methods like FedAvg and FedProx, as well as existing domain-agnostic CLIP-based approaches such as PromptFL and FedCLIP, our ADAPT consistently outperforms them across three benchmark scenarios.

\bibliography{egbib}
\clearpage
\section{Appendix}

\begin{algorithm}[ht!]
\caption{Training Process of ADAPT}
\scriptsize
\begin{algorithmic}[1]

\Statex \textbf{Input:}
\Statex \hspace*{\algorithmicindent} CLIP vision encoder $\mF_V$, text encoder $\mF_T$
\Statex \hspace*{\algorithmicindent} $n$ local datasets, each $\mD_i=\left\{(\text{[image]}, \text{[class name]})_j \right\}_{j=1}^{J}$
\Statex \hspace*{\algorithmicindent} Total communication rounds $T$, momentum coefficient $\alpha$
\Statex \textbf{Initialization:}
\Statex \hspace*{\algorithmicindent} Randomly initialize text prompts $[\mP_T^1]^0, \ldots, [\mP_T^n]^0$
\Statex \hspace*{\algorithmicindent} Randomly initialize visual prompts $[\mV] = \{[\vv]_1, \ldots, [\vv]_n\}$
\Statex \hspace*{\algorithmicindent} Broadcast the pretrained model and prompts to $n$ clients
\For {$t = 1$ to $T$}
    \State \textit{\# Local training in parallel}
    \For {$i = 1$ to $n$}
        \State Keep $\mF_V$ and $\mF_T$ frozen
        \For{$j=1$ to $J$}
            \State Compute $\vf_T^k = \mF_T(\mP_T^k, \text{[class name]}_j)$ for $k \in \{1, \ldots, n\}$
            \State Compute $\vf_V = \mF_V(\text{[cls]}, [\vv]_1, \ldots, [\vv]_n, \text{[image]}_j)$
            \State Extract attention scores $\vw = [w_1, \ldots, w_n]$ from $\mF_V$ using Eq.5
            \State Weighted sum: $\vf_T = \sum_{k=1}^n w_k \vf_T^k$
            \State Compute L$_2$ loss: $\gL=<\vf_V, \vf_T> / ||\vf_V||\cdot ||\vf_T||$
            \State Update $[\vv]_1, \ldots, [\vv]_n$ and $\mP_T^i$ by $\gL$
            \State Update $\mP_T^k, k\in\{1, \ldots, n\}, k\neq i$ by momentum: $\mP_T^k = \alpha\mP_T^k + (1-\alpha)[\mP_T^k]^{t-1}$
        \EndFor
    \EndFor

    \State \textit{\# Global aggregation in the server}
    
    \State Average $[\mV] = \frac{1}{n} \sum_{k=1}^n [\mV]^k$, where $[\mV]^k=\{[\vv]_1, \ldots, [\vv]_n\}$ obtained from \#$k$ client
    \State Assign $[\mP_T^k]^t=\mP_T^k$, where $\mP_T^k$ obtained from \#$k$ client
    \State Broadcast $[\mV], [\mP_T^k]^t (k\in\{1,\ldots,n\})$ to all clients
\EndFor
\end{algorithmic}
\label{algo:dpt}
\end{algorithm}

\textbf{Datasets}. We evaluate our ADAPT and baseline methods on the following three domain adaptation image classification benchmarks:
\begin{itemize}
    \item DomainNet~\cite{domainnet}. The DomainNet dataset has around 600,000 images spanning 345 categories from six domains, which covers diverse image styles including clipart, infograph, painting, quickdraw, real, and sketch. 
    \item OfficeHome~\cite{officehome}. The OfficeHome dataset consists of approximately 15,500 images depicting everyday objects in 65 classes. It further categorizes the images into four domains: art, clipart, product, and real-world.
    \item PACS~\cite{pacs}. The PACS dataset contains around 10,000 images drawn from seven categories and four domains, including photo, sketch, cartoon, and painting styles.
\end{itemize}

\begin{figure}
    \centering
    \includegraphics[width=0.45\textwidth]{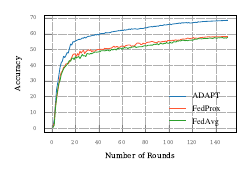}
    \caption{Comparison of performance to fine-tuning protocols on DomainNet dataset}
    \label{fig:compare3}
    \vspace{-10pt}
\end{figure}

\textbf{Communication Costs.} ADAPT markedly reduces communication overhead in federated learning by only transferring domain prompts, contrary to standard methods that share all trainable parameters. To provide a clear comparison, we have included the following results in Table~\ref{tab:compare2} and Figure~\ref{fig:compare3}. An additional benefit of this approach is its ability to produce favorable results without requiring a substantial volume of training data. As shown in Table 8(in the supplementary material), we obtain very competitive few-shot results by our prompt tuning technique. In practice, We avoid fine-tuning the CLIP model to maintain its visual-language alignment. Fine-tuning large models such as CLIP escalates communication expenses and impedes the rate of convergence. With an equivalent number of training iterations, the fine-tuning protocol often falls short to prompt learning. 

\begin{table}[H]
    \centering
    \caption{Comparison of Parameters and accuracy (\%) to fine-tuning protocols on DomainNet dataset. Our results are marked in \colorbox{cyan!10}{blue}. The best results in each domain are \textbf{bolded}} 
    \begin{tabular}{l>{\centering\arraybackslash}p{0.15\linewidth} >
    {\centering\arraybackslash}p{0.1\linewidth} 
    }
    \toprule
    Method & { Learnable params} & {acc.}  \\ 
    \midrule
    FedAvg~\cite{fedavg}   & 86M   &  57.6 \\
    FedProx~\cite{fedprox} &  86M     &  58.1 \\\rowcolor{cyan!5}
    ADAPT (ours)         & \textbf{16.9k} &  \textbf{68.4}   \\
    \bottomrule
    \end{tabular}
    \label{tab:compare2}
\end{table}

\begin{table*}[!t]
    \centering
    \caption{Few-shot accuracy (\%) on DomainNet. $n$-shot denotes training with $n$ samples per class and per domain. Our results are marked in \colorbox{cyan!10}{blue}. The best results are {\bf bolded}.}
    \begin{tabular}{lcccccccc}
    \toprule
    \bf Method & CLIP-based & full & 1-shot & 2-shot & 4-shot & 8-shot & 16-shot\\
    \midrule
    Single Domain Tuning & \cmark & 59.1 & 51.1 & 51.8 & 53.2 & 54.7 & 56.2 \\
    FedAvg (\textit{ResNet-50}) & \xmark & 54.7 & - & - & - & - & 15.1 \\
    FedAvg (\textit{ViT-Base/16}) & \xmark & 55.2 & - & - & - & - & 19.7 \\
    PromptFL & \cmark & 63.2 & 51.4 & 51.8 & 55.2 & 57.6 & 61.2\\
    FedCLIP &\cmark & 60.3 & 50.8 & 51.2 & 52.1 & 53.4 & 54.6\\
    \midrule
    \rowcolor{cyan!5}
    ADAPT (ours) & \cmark &  \textbf{68.4} & \textbf{55.4} & \textbf{57.2} & \textbf{60.3} & \textbf{62.7} &\textbf{64.5}\\\bottomrule
    \end{tabular}
    \label{tab:abl-fs}
\end{table*}

\begin{table*}
    \centering
    \caption{Test accuracy (\%) on DomainNet with 30 clients. Our results are marked in \colorbox{cyan!10}{blue}. The best results in each domain are \textbf{bolded}.}
    \begin{tabular}{l >{\centering\arraybackslash}p{0.1\linewidth} >{\centering\arraybackslash}p{0.08\linewidth} >{\centering\arraybackslash}p{0.08\linewidth} >{\centering\arraybackslash}p{0.08\linewidth} >{\centering\arraybackslash}p{0.08\linewidth} >{\centering\arraybackslash}p{0.08\linewidth} >{\centering\arraybackslash}p{0.08\linewidth}}
        \toprule
        \multirow{2}{*}{\textbf{Method}} & \multicolumn{7}{c}{\bf DomainNet}   \\
        \cmidrule(lr){2-8}
        & clipart   & infograph  & painting  & quickdraw  & real  & sketch  & average \\
        \midrule
        
        Zero-Shot CLIP & 66.1 & 40.6 & 62.3 & 13.5  & 80.4 & 58.5 & 53.6\\
        FedAvg & 37.6 & 56.4 & 55.6 & 31.0 & 71.9 & 57.2 & 51.6\\
        FedProx & 38.4 & 57.2 & 54.9 & 32.5 & 72.8 & 58.5 & 52.4\\
        PromptFL & 73.2 & 48.1 & 68.7 & 31.9 & 78.6 & 64.7 & 60.9\\
        FedCLIP & 72.7 & 47.0 & 66.2 & 32.8 & 76.9 & 57.2 & 58.8 \\\rowcolor{cyan!5}
    
        ADAPT (ours) & \textbf{75.8}   & \textbf{62.3}  &\textbf{69.0}  &\textbf{39.5}  &\textbf{83.9}  &\textbf{70.6}  &\textbf{66.9}\\
        
        \bottomrule
    \end{tabular}
    \label{tab:abl-dec}
\end{table*}

\noindent\textbf{Robustness to few-shot learning.} One of the primary advantages of prompt learning is the robustness to few-shot scenarios. We investigate if our dual prompt tuning method retains this merit in the context of federated learning. Therefore, we conduct few-shot learning experiments on DomainNet, employing 1, 2, 4, 8, and 16 training samples per category and per domain. We evaluate the other CLIP-based methods with the same setting, yet only test 16-shot performance for FedAvg as it fails to yield reasonable results with fewer training samples. The corresponding results are summarized in Table~\ref{tab:abl-fs}. As is shown, CLIP-based methods exhibit superior robustness against few-shot learning than FedAvg, which again demonstrates the significant benefits of using parameter-efficient approaches. Also, our ADAPT consistently outperforms the baselines in few-shot learning.



\textbf{Decentralization}. By default, we consider each domain in the dataset as a single client, leading to non-identical feature distributions yet the same class distribution across clients. To further testify our method's effectiveness and flexibility, we conduct a more challenging scenario on DomainNet where each domain is further divided into five clients by Dirichlet sampling, leading to 30 sub-datasets with either non-i.i.d. features or non-i.i.d. categories. Under this setup, we average the text prompt tokens for clients in the same domain at the aggregation step. The results are summarized in Table~\ref{tab:abl-dec}. Compared to our default setting which each domain is considered as one client, our ADAPT only has 1.5\% accuracy decrease when the dataset is further divided. In contrast, the conventional methods FedAvg and FedProx perform more sensitive to the non-i.i.d categories, with 3.6\% and 2.9\% accuracy decrease, respectively.

\end{document}